\documentclass[11pt]{article}
\usepackage{acl2002,url,alltt,epsfig}

\def\arXivhack{\vspace{-6pt}}

\title{NLTK: The Natural Language Toolkit}
\author{
Edward Loper and Steven Bird\\
Department of Computer and Information Science \\
University of Pennsylvania, Philadelphia, PA 19104-6389, USA
}
\date{}

\newenvironment{sv}{\small\begin{alltt}}{\end{alltt}\normalsize}

\begin{document}
\maketitle

\begin{abstract}
NLTK, the Natural Language Toolkit, is a suite of open source program modules,
tutorials and problem sets, providing ready-to-use computational
linguistics courseware.  NLTK covers symbolic and statistical natural
language processing, and is interfaced to annotated corpora.  Students
augment and replace existing components, learn structured
programming by example, and manipulate sophisticated models from the
outset.
\end{abstract}

\section{Introduction}

Teachers of introductory courses on computational linguistics are
often faced with the challenge of setting up a practical programming
component for student assignments and projects.  This is a difficult
task because different computational linguistics domains require a
variety of different data structures and functions, and because a
diverse range of topics may need to be included in the syllabus.

A widespread practice is to employ multiple programming languages,
where each language provides native data structures and functions that
are a good fit for the task at hand.  For example, a course might use
Prolog for parsing, Perl for corpus processing, and a finite-state
toolkit for morphological analysis.  By relying on the built-in
features of various languages, the teacher avoids having to develop a
lot of software infrastructure.

An unfortunate consequence is that a significant part of such courses
must be devoted to teaching programming languages.  Further, many
interesting projects span a variety of domains, and would require that
multiple languages be bridged.  For example, a student project that
involved syntactic parsing of corpus data from a morphologically rich
language might involve all three of the languages mentioned above:
Perl for string processing; a finite state toolkit for morphological
analysis; and Prolog for parsing.
It is clear that these considerable overheads and shortcomings warrant
a fresh approach.

Apart from the practical component, computational linguistics courses
may also depend on software for in-class demonstrations.  This context
calls for highly interactive graphical user interfaces, making it
possible to view program state (e.g. the chart of a chart parser),
observe program execution step-by-step (e.g. execution of a
finite-state machine), and even make minor modifications to programs
in response to ``what if'' questions from the class.  Because of these
difficulties it is common to avoid live demonstrations, and keep
classes for theoretical presentations only.  Apart from being dull,
this approach leaves students to solve important practical problems on
their own, or to deal with them less efficiently in office hours.

In this paper we introduce a new approach to the above challenges, a
streamlined and flexible way of organizing the practical component of
an introductory computational linguistics course.  We describe NLTK,
the Natural Language Toolkit, which we have developed in conjunction
with a course we have taught at the University of Pennsylvania.

The Natural Language Toolkit is available under an open source license
from \mbox{\url{http://nltk.sf.net/}}.  NLTK runs on all platforms
supported by Python, including Windows, OS X, Linux, and Unix.
\arXivhack

\section{Choice of Programming Language}
\label{sec:python}

The most basic step in setting up a practical component is choosing a
suitable programming language.  A number of considerations influenced
our choice.  First, the language must have a shallow learning curve,
so that novice programmers get immediate rewards for their efforts.
Second, the language must support rapid prototyping and a short
develop/test cycle; an obligatory compilation step is a serious
detraction.  Third, the code should be self-documenting, with a
transparent syntax and semantics.  Fourth, it should be easy to write
structured programs, ideally object-oriented but without the burden
associated with languages like C++.  Finally, the language must have
an easy-to-use graphics library to support the development of
graphical user interfaces.

In surveying the available languages, we believe that Python offers an
especially good fit to the above requirements.  Python is an
object-oriented scripting language developed by Guido van Rossum
and available on all platforms (\url{www.python.org}).  Python offers a
shallow learning curve; it was designed to be easily learnt by
children \cite{rossum99}.  As an interpreted language, Python is
suitable for rapid prototyping.  Python code is exceptionally
readable, and it has been praised as ``executable pseudocode.''
Python is an object-oriented language, but not punitively so, and it
is easy to encapsulate data and methods inside Python classes.
Finally, Python has an interface to the Tk graphics toolkit
\cite{tkinter}, and writing graphical interfaces is straightforward.
\arXivhack

\section{Design Criteria}
\label{sec:criteria}

Several criteria were considered in the design and implementation of
the toolkit.  These design criteria are listed in the order of their
importance.  It was also important to decide what goals the toolkit
would \emph{not} attempt to accomplish; we therefore include an
explicit set of non-requirements, which the toolkit is not expected to
satisfy.

\subsection{Requirements}

\paragraph{\textit{Ease of Use.}} The primary purpose of the toolkit is
to allow students to concentrate on building natural language
processing (NLP) systems.  The more time students must spend learning
to use the toolkit, the less useful it is.

\paragraph{\textit{Consistency.}} The toolkit should use consistent data
structures and interfaces.

\paragraph{\textit{Extensibility.}} The toolkit should easily
accommodate new components, whether those components replicate or
extend the toolkit's existing functionality.  The toolkit should be
structured in such a way that it is obvious where new extensions would
fit into the toolkit's infrastructure.

\paragraph{\textit{Documentation.}} The toolkit, its data structures,
and its implementation all need to be carefully and thoroughly
documented.  All nomenclature must be carefully chosen and
consistently used.

\paragraph{\textit{Simplicity.}} The toolkit should structure the
complexities of building NLP systems, not hide them.  Therefore, each
class defined by the toolkit should be simple enough that a student
could implement it by the time they finish an introductory course in
computational linguistics.

\paragraph{\textit{Modularity.}} The interaction between different
components of the toolkit should be kept to a minimum, using simple,
well-defined interfaces.  In particular, it should be possible to
complete individual projects using small parts of the toolkit, without
worrying about how they interact with the rest of the toolkit.  This
allows students to learn how to use the toolkit incrementally
throughout a course.  Modularity also makes it easier to change and
extend the toolkit.

\subsection{Non-Requirements}

\paragraph{\textit{Comprehensiveness.}} The toolkit is not intended to
provide a comprehensive set of tools.  Indeed, there should be a wide
variety of ways in which students can extend the toolkit.

\paragraph{\textit{Efficiency.}} The toolkit does not need to be highly
optimized for runtime performance.  However, it should be efficient
enough that students can use their NLP systems to perform real tasks.

\paragraph{\textit{Cleverness.}} Clear designs and implementations are
far preferable to ingenious yet indecipherable ones.
\arXivhack

\section{Modules}
\label{sec:modules}

The toolkit is implemented as a collection of independent
\emph{modules}, each of which defines a specific data structure or
task.

A set of core modules defines basic data types and processing systems
that are used throughout the toolkit.  The \texttt{token} module
provides basic classes for processing individual elements of text,
such as words or sentences.  The \texttt{tree} module defines data
structures for representing tree structures over text, such as syntax
trees and morphological trees.  The \texttt{probability} module
implements classes that encode frequency distributions and probability
distributions, including a variety of statistical smoothing
techniques.

The remaining modules define data structures and interfaces for
performing specific NLP tasks.  This list of modules will grow over
time, as we add new tasks and algorithms to the toolkit.

\subsection*{Parsing Modules}

The \texttt{parser} module defines a high-level interface for
producing trees that represent the structures of texts.  The
\texttt{chunkparser} module defines a sub-interface for parsers that
identify non-overlapping linguistic groups (such as base noun phrases)
in unrestricted text.

Four modules provide implementations for these abstract interfaces.
The \texttt{srparser} module implements a simple shift-reduce parser.
The \texttt{chartparser} module defines a flexible parser that uses a
\emph{chart} to record hypotheses about syntactic constituents.  The
\texttt{pcfgparser} module provides a variety of different parsers for
probabilistic grammars.  And the \texttt{rechunkparser} module defines
a transformational regular-expression based implementation of the
chunk parser interface.

\subsection*{Tagging Modules}

The \texttt{tagger} module defines a standard interface for augmenting
each token of a text with supplementary information, such as its part
of speech or its WordNet synset tag; and provides several different
implementations for this interface.

\subsection*{Finite State Automata}

The \texttt{fsa} module defines a data type for encoding finite state
automata; and an interface for creating automata from regular
expressions.

\subsection*{Type Checking}

Debugging time is an important factor in the toolkit's ease of use.
To reduce the amount of time students must spend debugging their code,
we provide a type checking module, which can be used to ensure that
functions are given valid arguments.  The type checking module is
used by all of the basic data types and processing classes.

Since type checking is done explicitly, it can slow the toolkit down.
However, when efficiency is an issue, type checking can be easily
turned off; and with type checking is disabled, there is no
performance penalty.

\subsection*{Visualization}

Visualization modules define graphical interfaces for viewing and
manipulating data structures, and graphical tools for experimenting
with NLP tasks.  The \texttt{draw.tree} module provides a simple
graphical interface for displaying tree structures.  The
\texttt{draw.tree\_edit} module provides an interface for building and
modifying tree structures.  The \texttt{draw.plot\_graph} module can be
used to graph mathematical functions.  The \texttt{draw.fsa} module
provides a graphical tool for displaying and simulating finite state
automata.  The \texttt{draw.chart} module provides an interactive
graphical tool for experimenting with chart parsers.

The visualization modules provide interfaces for interaction and
experimentation; they do not directly implement NLP data structures or
tasks.  Simplicity of implementation is therefore less of an issue for
the visualization modules than it is for the rest of the toolkit.

\subsection*{Text Classification}

The \texttt{classifier} module defines a standard interface for
classifying texts into categories.  This interface is currently
implemented by two modules.  The \texttt{classifier.naivebayes} module
defines a text classifier based on the Naive Bayes assumption. The
\texttt{classifier.maxent} module defines the maximum entropy model
for text classification, and implements two algorithms for training
the model: Generalized Iterative Scaling and Improved Iterative
Scaling.

The \texttt{classifier.feature} module provides a standard encoding
for the information that is used to make decisions for a particular
classification task.  This standard encoding allows students to
experiment with the differences between different text classification
algorithms, using identical feature sets.

The \texttt{classifier.featureselection} module defines a standard
interface for choosing which features are relevant for a particular
classification task.  Good feature selection can significantly improve
classification performance.
\arXivhack

\section{Documentation}
\label{sec:documentation}

The toolkit is accompanied by extensive documentation that explains
the toolkit, and describes how to use and extend it.  This
documentation is divided into three primary categories:

\paragraph{\textit{Tutorials}} teach students how to use the toolkit,
in the context of performing specific tasks.  Each tutorial focuses on
a single domain, such as tagging, probabilistic systems, or text
classification.  The tutorials include a high-level discussion that
explains and motivates the domain, followed by a detailed
walk-through that uses examples to show how NLTK can be used to
perform specific tasks.

\paragraph{\textit{Reference Documentation}} provides precise
definitions for every module, interface, class, method, function, and
variable in the toolkit.  It is automatically extracted from docstring
comments in the Python source code, using Epydoc \cite{epydoc}.

\paragraph{\textit{Technical Reports}} explain and justify the
toolkit's design and implementation.  They are used by the developers
of the toolkit to guide and document the toolkit's construction.
Students can also consult these reports if they would like further
information about how the toolkit is designed, and why it is designed
that way.
\arXivhack

\section{Uses of NLTK}
\label{sec:uses}

\subsection{Assignments}

NLTK can be used to create student assignments of varying difficulty
and scope.  
In the simplest assignments, students experiment with an existing
module.  The wide variety of existing modules provide many opportunities
for creating these simple assignments.
Once students become more familiar with the toolkit, they can be asked
to make minor changes or extensions to an existing module.
A more challenging task is to develop a new module.  Here, NLTK
provides some useful starting points: predefined interfaces and data
structures, and existing modules that implement the same interface.

\subsubsection*{Example: Chunk Parsing}

As an example of a moderately difficult assignment, we asked students
to construct a chunk parser that correctly identifies base noun phrase
chunks in a given text, by defining a cascade of transformational
chunking rules.  The NLTK \texttt{rechunkparser} module provides a
variety of regular-expression based rule types, which the students can
instantiate to construct complete rules.  For example,
\texttt{ChunkRule('<NN.*>')} builds chunks from sequences of
consecutive nouns; \texttt{ChinkRule('<VB.>')} excises verbs from
existing chunks; \texttt{SplitRule('<NN>', '<DT>')} splits any
existing chunk that contains a singular noun followed by determiner
into two pieces; and \texttt{MergeRule('<JJ>', '<JJ>')} combines two
adjacent chunks where the first chunk ends and the second chunk starts
with adjectives.

The chunking tutorial motivates chunk parsing, describes each rule
type, and provides all the necessary code for the assignment.  The
provided code is responsible for loading the chunked, part-of-speech
tagged text using an existing tokenizer, creating an unchunked version
of the text, applying the chunk rules to the unchunked text, and
scoring the result.  Students focus on the NLP task only -- providing
a rule set with the best coverage.

In the remainder of this section we reproduce some of the cascades
created by the students.  The first example illustrates a combination
of several rule types:

\begin{sv}
cascade = [
  ChunkRule('<DT><NN.*><VB.><NN.*>'),
  ChunkRule('<DT><VB.><NN.*>'),
  ChunkRule('<.*>'),
  UnChunkRule('<IN|VB.*|CC|MD|RB.*>'),
  UnChunkRule("<,|{\textbackslash}{\textbackslash}.|``|''>"),
  MergeRule('<NN.*|DT|JJ.*|CD>',
            '<NN.*|DT|JJ.*|CD>'),
  SplitRule('<NN.*>', '<DT|JJ>')
]
\end{sv}

The next example illustrates a brute-force statistical approach.  The
student calculated how often each part-of-speech tag was included in a
noun phrase.  They then constructed chunks from any sequence of tags
that occurred in a noun phrase more than 50\% of the time.

\begin{sv}
cascade = [
  ChunkRule('<{\textbackslash}{\textbackslash}\$|CD|DT|EX|PDT
             |PRP.*|WP.*|{\textbackslash}{\textbackslash}\#|FW
             |JJ.*|NN.*|POS|RBS|WDT>*')
]
\end{sv}

In the third example, the student constructed a single chunk
containing the entire text, and then excised all elements that did not
belong.

\begin{sv}
cascade = [
  ChunkRule('<.*>+')
  ChinkRule('<VB.*|IN|CC|R.*|MD|WRB|TO|.|,>+')
]
\end{sv}

\subsection{Class demonstrations}

NLTK provides graphical tools that can be used in class demonstrations
to help explain basic NLP concepts and algorithms.  These interactive
tools can be used to display relevant data structures and to show the
step-by-step execution of algorithms.  Both data structures and
control flow can be easily modified during the demonstration, in
response to questions from the class.

Since these graphical tools are included with the toolkit, they can
also be used by students.  This allows students to experiment at home
with the algorithms that they have seen presented in class.

\subsubsection*{Example: The Chart Parsing Tool}

The chart parsing tool is an example of a graphical tool provided by
NLTK.  This tool can be used to explain the basic concepts behind
chart parsing, and to show how the algorithm works.  Chart parsing is
a flexible parsing algorithm that uses a data structure called a
\emph{chart} to record hypotheses about syntactic constituents.  Each
hypothesis is represented by a single \emph{edge} on the chart.  A set
of \emph{rules} determine when new edges can be added to the chart.
This set of rules controls the overall behavior of the parser (e.g.,
whether it parses top-down or bottom-up).

\begin{figure}
\centerline{\epsfig{figure=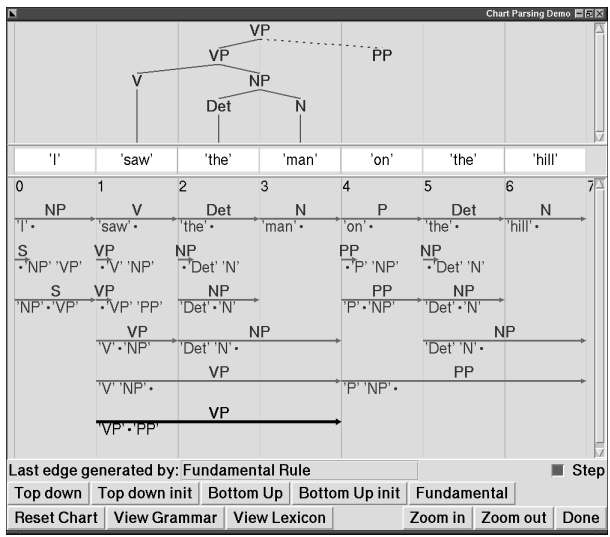,width=\linewidth}}
\caption{Chart Parsing Tool}\label{fig:chartparse}
\vspace*{2ex}\hrule
\end{figure}

The chart parsing tool demonstrates the process of parsing a single
sentence, with a given grammar and lexicon.  Its display is divided
into three sections: the bottom section displays the chart; the middle
section displays the sentence; and the top section displays the
partial syntax tree corresponding to the selected edge.  Buttons along
the bottom of the window are used to control the execution of the
algorithm.  The main display window for the chart parsing tool is
shown in Figure~\ref{fig:chartparse}.   

This tool can be used to explain several different aspects of chart
parsing.  First, it can be used to explain the basic chart data
structure, and to show how edges can represent hypotheses about
syntactic constituents.  It can then be used to demonstrate and
explain the individual rules that the chart parser uses to create new
edges.  Finally, it can be used to show how these individual rules
combine to find a complete parse for a given sentence.

To reduce the overhead of setting up demonstrations during lecture,
the user can define a list of preset charts.  The tool can then be
reset to any one of these charts at any time.

The chart parsing tool allows for flexible control of the parsing
algorithm.  At each step of the algorithm, the user can select which
rule or strategy they wish to apply.  This allows the user to
experiment with mixing different strategies (e.g., top-down and
bottom-up).  The user can exercise fine-grained control over the
algorithm by selecting which edge they wish to apply a rule to.  This
flexibility allows lecturers to use the tool to respond to a wide
variety of questions; and allows students to experiment with different
variations on the chart parsing algorithm.

\subsection{Advanced Projects}

NLTK provides students with a flexible framework for advanced
projects.  Typical projects involve the development of entirely new
functionality for a previously unsupported NLP task, or the
development of a complete system out of existing and new modules.

The toolkit's broad coverage allows students to explore a wide variety
of topics.  In our introductory computational linguistics course,
topics for student projects included text generation, word sense
disambiguation, collocation analysis, and morphological analysis.

NLTK eliminates the tedious infrastructure-building that is typically
associated with advanced student projects by providing students with
the basic data structures, tools, and interfaces that they need.  This
allows the students to concentrate on the problems that interest them.

The collaborative, open-source nature of the toolkit can provide
students with a sense that their projects are meaningful
contributions, and not just exercises.  Several of the students in our
course have expressed interest in incorporating their projects into
the toolkit.

Finally, many of the modules included in the toolkit provide students
with good examples of what projects should look like, with well
thought-out interfaces, clean code structure, and thorough
documentation.

\subsubsection*{Example: Probabilistic Parsing}

The probabilistic parsing module was created as a class project for a
statistical NLP course.  The toolkit provided the basic data types and
interfaces for parsing.  The project extended these, adding a new
probabilistic parsing interface, and using subclasses to create a
probabilistic version of the context free grammar data structure.
These new components were used in conjunction with several existing
components, such as the chart data structure, to define two
implementations of the probabilistic parsing interface.  Finally, a
tutorial was written that explained the basic motivations and concepts
behind probabilistic parsing, and described the new interfaces, data
structures, and parsers.
\arXivhack

\section{Evaluation}
\label{sec:evaluation}

We used NLTK as a basis for the assignments and student projects in
CIS-530, an introductory computational linguistics class taught at the
University of Pennsylvania.  CIS-530 is a graduate level class,
although some advanced undergraduates were also enrolled.  Most
students had a background in either computer science or linguistics
(and occasionally both).  Students were required to complete five assignments,
two exams, and a final project.  All class materials are available
from the course website \mbox{\url{http://www.cis.upenn.edu/~cis530/}}.

The experience of using NLTK was very positive, both for us and for
the students.  The students liked the fact that they could do
interesting projects from the outset.  They also liked being able to
run everything on their computer at home.  The students found the
extensive documentation very helpful for learning to use the toolkit.
They found the interfaces defined by NLTK intuitive, and appreciated
the ease with which they could combine different components to create
complete NLP systems.

We did encounter a few difficulties during the semester.  One problem
was finding large clean corpora that the students could use for their
assignments.  Several of the students needed assistance finding
suitable corpora for their final projects.  Another issue was the fact
that we were actively developing NLTK during the semester; some
modules were only completed one or two weeks before the students used
them.  As a result, students who worked at home needed to download new
versions of the toolkit several times throughout the semester.
Luckily, Python has extensive support for installation scripts, which
made these upgrades simple.  The students encountered a couple
of bugs in the toolkit, but none were serious, and all were quickly
corrected.
\arXivhack

\section{Other Approaches}
\label{sec:approaches}

The computational component of computational linguistics courses takes
many forms.  In this section we briefly review a selection of approaches,
classified according to the (original) target audience.

{\bf Linguistics Students.}
Various books introduce programming or computing to linguists.
These are elementary on the computational side, providing a
gentle introduction to students having no prior experience
in computer science.  Examples of such books are:
\emph{Using Computers in Linguistics}
\cite{Lawler98}, and
\emph{Programming for Linguistics: Java Technology for Language
Researchers} \cite{Hammond02}.

{\bf Grammar Developers.}
Infrastructure for grammar development has a long history in
unification-based (or constraint-based) grammar frameworks, from DCG
\cite{PereiraWarren80} to HPSG \cite{PollardSag94}.  Recent work includes
\cite{Copestake00,Baldridge02}.  A concurrent development has been the
finite state toolkits, such as the Xerox toolkit \cite{Beesley02}.  This
work has found widespread pedagogical application.

{\bf Other Researchers and Developers.}
A variety of toolkits have been created for research or R\&D
purposes.  Examples include
the \emph{CMU-Cambridge Statistical Language Modeling Toolkit}
\cite{Clarkson97},
the \emph{EMU Speech Database System} \cite{Harrington99},
the \emph{General Architecture for Text Engineering} \cite{Bontcheva02},
the \emph{Maxent Package for Maximum Entropy Models} \cite{maxent},
and the \emph{Annotation Graph Toolkit} \cite{MaedaBird02}.
Although not originally motivated by pedagogical needs, all of these
toolkits have pedagogical applications and many have already been
used in teaching.
\arXivhack

\section{Conclusions and Future Work}
\label{sec:conclusion}

NLTK provides a simple, extensible, uniform framework for assignments,
projects, and class demonstrations.  It is well documented, easy to
learn, and simple to use.  We hope that NLTK will allow computational
linguistics classes to include more hands-on experience with using and
building NLP components and systems.

NLTK is unique in its combination of three factors.
First, it was deliberately designed as courseware and gives pedagogical
goals primary status.  Second, its target audience consists of both
linguists and computer scientists, and it is accessible and challenging
at many levels of prior computational skill.  Finally, it is based on
an object-oriented scripting language supporting rapid prototyping and
literate programming.

We plan to continue extending the breadth of materials covered by
the toolkit.  We are currently working on NLTK modules for Hidden
Markov Models, language modeling, and tree adjoining grammars.  We
also plan to increase the number of algorithms implemented by some
existing modules, such as the text classification module.

Finding suitable corpora is a prerequisite for many student assignments and
projects.  We are therefore putting together a collection of corpora
containing data appropriate for every module defined by the toolkit.

NLTK is an open source project, and we welcome any contributions.
Readers who are interested in contributing to NLTK, or who have
suggestions for improvements, are encouraged to contact the authors.
\arXivhack

\section{Acknowledgments}
\label{sec:acknowledgments}

We are indebted to our students for feedback on the toolkit, and to
anonymous reviewers, Jee Bang, and the workshop organizers for
comments on an earlier version of this paper.  We are grateful to
Mitch Marcus and the Department of Computer and Information Science at
the University of Pennsylvania for sponsoring the work reported here.

\bibliographystyle{acl}

\begin{thebibliography}{}

\bibitem[\protect\citename{Baldridge \bgroup et al.\egroup }2002a]{Baldridge02}
Jason Baldridge, John Dowding, and Susana Early.
\newblock 2002a.
\newblock Leo: an architecture for sharing resources for unification-based
  grammars.
\newblock In {\em Proceedings of the Third Language Resources and Evaluation
  Conference}. Paris: European Language Resources Association.
\newblock \\\url{http://www.iccs.informatics.ed.ac.uk/~jmb/leo-lrec.ps.gz}.

\bibitem[\protect\citename{Baldridge \bgroup et al.\egroup }2002b]{maxent}
Jason Baldridge, Thomas Morton, and Gann Bierner.
\newblock 2002b.
\newblock The {MaxEnt} project.
\newblock \\\url{http://maxent.sourceforge.net/}.

\bibitem[\protect\citename{Beesley and Karttunen}2002]{Beesley02}
Kenneth~R. Beesley and Lauri Karttunen.
\newblock 2002.
\newblock {\em Finite-State Morphology: Xerox Tools and Techniques}.
\newblock Studies in Natural Language Processing. Cambridge University Press.

\bibitem[\protect\citename{Bontcheva \bgroup et al.\egroup }2002]{Bontcheva02}
Kalina Bontcheva, Hamish Cunningham, Valentin Tablan, Diana Maynard, and Oana
  Hamza.
\newblock 2002.
\newblock Using {GATE} as an environment for teaching {NLP}.
\newblock In {\em Proceedings of the ACL Workshop on Effective Tools and
  Methodologies for Teaching NLP and CL}. Somerset, NJ: Association for
  Computational Linguistics.

\bibitem[\protect\citename{Clarkson and Rosenfeld}1997]{Clarkson97}
Philip~R. Clarkson and Ronald Rosenfeld.
\newblock 1997.
\newblock Statistical language modeling using the {CMU-Cambridge Toolkit}.
\newblock In {\em Proceedings of the 5th European Conference on Speech
  Communication and Technology (EUROSPEECH '97)}.
\newblock \url{http://svr-www.eng.cam.ac.uk/~prc14/eurospeech97.ps}.

\bibitem[\protect\citename{Copestake}2000]{Copestake00}
Ann Copestake.
\newblock 2000.
\newblock The (new) {LKB} system.
\newblock \\\url{http://www-csli.stanford.edu/~aac/doc5-2.pdf}.

\bibitem[\protect\citename{Hammond}2002]{Hammond02}
Michael Hammond.
\newblock 2002.
\newblock {\em Programming for Linguistics: Java Technology for Language
  Researchers}.
\newblock Oxford: Blackwell.
\newblock In press.

\bibitem[\protect\citename{Harrington and Cassidy}1999]{Harrington99}
Jonathan Harrington and Steve Cassidy.
\newblock 1999.
\newblock {\em Techniques in Speech Acoustics}.
\newblock Kluwer.

\bibitem[\protect\citename{Lawler and Dry}1998]{Lawler98}
John~M. Lawler and Helen~Aristar Dry, editors.
\newblock 1998.
\newblock {\em Using Computers in Linguistics}.
\newblock London: Routledge.

\bibitem[\protect\citename{Loper}2002]{epydoc}
Edward Loper.
\newblock 2002.
\newblock Epydoc.
\newblock \\\url{http://epydoc.sourceforge.net/}.

\bibitem[\protect\citename{Lundh}1999]{tkinter}
Fredrik Lundh.
\newblock 1999.
\newblock An introduction to tkinter.
\newblock
  \\\url{http://www.pythonware.com/library/tkinter/introduction/index.htm}.

\bibitem[\protect\citename{Maeda \bgroup et al.\egroup }2002]{MaedaBird02}
Kazuaki Maeda, Steven Bird, Xiaoyi Ma, and Haejoong Lee.
\newblock 2002.
\newblock Creating annotation tools with the annotation graph toolkit.
\newblock In {\em Proceedings of the Third International Conference on Language
  Resources and Evaluation}.
\newblock \url{http://arXiv.org/abs/cs/0204005}.

\bibitem[\protect\citename{Pereira and Warren}1980]{PereiraWarren80}
Fernando C.~N. Pereira and David H.~D. Warren.
\newblock 1980.
\newblock Definite clause grammars for language analysis -- a survey of the
  formalism and a comparison with augmented transition grammars.
\newblock {\em Artificial Intelligence}, 13:231--78.

\bibitem[\protect\citename{Pollard and Sag}1994]{PollardSag94}
Carl Pollard and Ivan~A. Sag.
\newblock 1994.
\newblock {\em Head-Driven Phrase Structure Grammar}.
\newblock Chicago University Press.

\bibitem[\protect\citename{van Rossum}1999]{rossum99}
Guido van Rossum.
\newblock 1999.
\newblock Computer programming for everybody.
\newblock Technical report, Corporation for National Research Initiatives.
\newblock \url{http://www.python.org/doc/essays/cp4e.html}.

\end{thebibliography}

\end{document}